\documentclass[sigconf]{aamas}

\usepackage{balance} 

\usepackage{multirow}
\usepackage{algorithm}
\usepackage{algorithmic}
\usepackage{enumitem}
\usepackage{url}
\usepackage{color}
\usepackage{booktabs}
\usepackage{makecell}
\usepackage{tabularx}
\usepackage{hyperref}
\usepackage{multicol}

\newcommand{\cmmnt}[1]{\ignorespaces}
\newcommand{\sllm}{SySLLM}
\newcommand{\hili}{HIGHLIGHTS}

\setcopyright{ifaamas}
\acmConference[AAMAS '26]{Proc.\@ of the 25th International Conference
on Autonomous Agents and Multiagent Systems (AAMAS 2026)}{May 25 -- 29, 2026}
{Paphos, Cyprus}{C.~Amato, L.~Dennis, V.~Mascardi, J.~Thangarajah (eds.)}
\copyrightyear{2026}
\acmYear{2026}
\acmDOI{}
\acmPrice{}
\acmISBN{}

\acmSubmissionID{370}

\title[AAMAS-2026 Formatting Instructions]{From Actions to Words: Towards Abstractive-Textual Policy Summarization in RL}

\author{Sahar Admoni}
\affiliation{
  \institution{Technion -- IIT}
  \city{Haifa}
  \country{Israel}
  }
\email{saharad@campus.technion.ac.il}

\author{Assaf Hallak}
\affiliation{
  \institution{Nvidia Research}
  \city{Tel Aviv}
  \country{Israel}
  }
\email{ahallak@nvidia.com}

\author{Yftah Ziser}
\affiliation{
  \institution{Nvidia Research}
  \city{Tel Aviv}
  \country{Israel}
  }
\email{yziser@nvidia.com}

\author{Omer Ben-Porat}
\affiliation{
  \institution{Technion -- IIT}
  \city{Haifa}
  \country{Israel}
  }
\email{omerbp@technion.ac.il}

\author{Ofra Amir}
\affiliation{
  \institution{Technion -- IIT}
  \city{Haifa}
  \country{Israel}
  }
\email{oamir@technion.ac.il}

\begin{abstract}
Explaining reinforcement learning agents is challenging because policies emerge from complex reward structures and neural representations that are difficult for humans to interpret. Existing approaches often rely on curated demonstrations that expose local behaviors but provide limited insight into an agent’s global strategy, leaving users to infer intent from raw observations.
We propose \emph{SySLLM} (Synthesized Summary using Large Language Models), a framework that reframes policy interpretation as a language-generation problem. Instead of visual demonstrations, SySLLM converts spatiotemporal trajectories into structured text and prompts an LLM to generate coherent summaries describing the agent’s goals, exploration style, and decision patterns.
SySLLM scales to long-horizon, semantically rich environments without task-specific fine-tuning, leveraging LLM world knowledge and compositional reasoning to capture latent behavioral structure across policies. Expert evaluations show strong alignment with human analyses, and a large-scale user study found that 75.5\% of participants preferred SySLLM summaries over state-of-the-art demonstration-based explanations. Together, these results position \emph{abstractive textual} summarization as a paradigm for interpreting complex RL behavior.\footnote{Code available at \url{https://github.com/saharad1/SySLLM}}
\end{abstract}


\newcommand{\BibTeX}{\rm B\kern-.05em{\sc i\kern-.025em b}\kern-.08em\TeX}

\begin{document}


\pagestyle{fancy}
\fancyhead{}


\maketitle 


\section{Introduction}
Reinforcement learning (RL) agents are increasingly deployed in sequential decision-making domains, yet their policies remain opaque to human stakeholders. This opacity limits trust, adoption, and effective debugging. 
While many explainable reinforcement learning (XRL) methods provide \emph{local} insight into individual states or decisions, our focus is on \emph{global} explanations that aim to capture an agent's strategy across trajectories.
Existing paradigms face trade-offs between \emph{expressiveness}, \emph{scalability}, and \emph{faithfulness}. Extractive methods such as saliency maps~\cite{greydanus2018visualizing,puri2019explain,atrey2019exploratory,samadi2024safe} and demonstration-based policy summaries~\cite{amir2018agent,frost2022explaining,liu2023learning,deshmukh2023explaining} capture only fragments of behavior. Symbolic approaches such as rules or decision trees~\cite{sequeira2020interestingness,peng2022inherently,mccarthy2022boolean,parham2023explaining} often collapse in high-dimensional or partially observed environments. Across these approaches, a core flaw persists: they transform observations but do not synthesize them into holistic accounts, leaving users to reconstruct intent, adaptability, and failure modes from fragmented evidence~\cite{anderson2020mental, amitaisurvey}.  

Large language models (LLMs) appear to offer a promising alternative. Their capacity for abstraction, compositional reasoning, and natural language generation~\cite{zhao2023survey} suggests that they could distill trajectories into human-readable accounts of agent behavior. However, applying them to RL policies is far from straightforward. LLMs are trained on static text-based corpora, while RL agents generate dynamic spatio-temporal trajectories grounded in states, actions, and rewards~\cite{li2024advancing}. This mismatch in modality makes faithful policy summarization non-trivial. Specifically, naive application risks producing fluent and plausibly sounding behavior descriptions that do not accurately reflect the underlying policy.  

To address this challenge, we propose a \textit{textual-abstractive paradigm} for policy summarization, in which explanations are expressed as natural language narratives integrating evidence from a database of policy execution traces, which we term the textual experience buffer (TEB). Unlike extractive or symbolic methods, this paradigm explicitly targets policy-level regularities and behavioral motifs. We formalize the task as a mapping from an agent's experience buffer to a textual summary, guided by the desiderata of \emph{expressiveness}, \emph{scalability}, and \emph{faithfulness}. To realize these principles in practice, we adopt a conceptual optimization view in which summaries balance \emph{coverage}, \emph{parsimony}, and \emph{fidelity}. This formulation grounds both our methodology and evaluation, ensuring that system design choices are principled and directly address the unique challenges of applying LLMs to RL.  

Building on this formulation, we introduce \textbf{SySLLM}, a framework that leverages LLMs to generate global policy summaries. SySLLM operates in two stages: agent–environment trajectories are first transformed into structured natural language descriptions of observations and actions, which are then synthesized into higher-level accounts through carefully designed prompting. To scale beyond context window limitations, SySLLM performs hierarchical summarization over large buffers. To mitigate variability in LLM outputs, it generates multiple candidate summaries and aggregates them into a consensus using embedding-based similarity. Together, these mechanisms ensure that SySLLM produces summaries that are general enough to capture policy-level regularities while specific enough to reflect distinctive behaviors.

We evaluated SySLLM across five MiniGrid environments and the Crafter domain, covering nine qualitatively distinct agent policies. Expert evaluation shows a strong alignment between SySLLM summaries and expert summaries, achieving high recall and precision scores that reflect faithful coverage of expert-identified behaviors. A user study with 192 participants further demonstrates that users strongly prefer textual summaries over demonstration-based summaries such as HIGHLIGHTS-DIV~\cite{amir2018highlights}, while performing equally well or better on policy identification tasks.  

Our main contributions are threefold:  
(1) we introduce and formalize the task of \emph{abstractive-textual policy summarization} in RL, framing it as a mapping from an agent's experience buffer to natural language narratives defined by desiderata of expressiveness, scalability, and faithfulness;  
(2) we present \textbf{SySLLM}, a framework that leverages LLMs to synthesize structured trajectory descriptions into coherent global summaries, incorporating hierarchical summarization and consensus aggregation to address long horizons and variability in outputs, and 
(3) we provide extensive empirical validation through expert evaluations and a large-scale user study, demonstrating that SySLLM produces faithful summaries that are strongly preferred by participants over state-of-the-art demonstration-based baselines.

\section{Related Work}
Prior work in XRL spans both local explanations of individual decisions and global summaries of agent behavior, with the shared goal of improving policy interpretability for humans~\cite{milani2022survey}.
Saliency and visualization methods highlight influential inputs~\cite{greydanus2018visualizing,huber2022benchmarking} but are often local and fragile; demonstration-based methods summarize behavior via selected trajectories~\cite{amir2018agent,deshmukh2023explaining} but place interpretive burden on users; surrogate models distill policies into rules or decision trees~\cite{bastani2018verifiable,coppens2019distilling}, though with fidelity--scalability trade-offs; and causal or reward-based explanations~\cite{madumal2020explainable,juozapaitis2019explainable} provide structured insights but typically require access to internals or domain expertise. Collectively, these techniques emphasize fragments or simplified proxies rather than synthesizing global accounts of an agent’s strategy.

Large language models (LLMs) bring complementary capabilities of abstraction, reasoning, and fluent text generation~\cite{brown2020language}. While prior work mainly uses LLMs during training---for example, guiding exploration or constructing world models~\cite{du2023guiding,guan2023leveraging}---their potential as \emph{explanation generators} has received less attention. Some studies prompt LLMs to narrate behavior in real time~\cite{wang2023voyager,yao2023react} or to build symbolic simulators~\cite{ahn2022can}, but these efforts are often ungrounded in actual dynamics and focus on local rather than global explanations.

Natural language explanations for RL agents have also been studied. The early template-based methods~\cite{hayes2017improving,peng2022inherently} prioritized accessibility, but were brittle, while neural rationalization approaches translated trajectories into free form text~\cite{ehsan2018rationalization,mccalmon2022caps}. These provide interpretability but generally rely on handcrafted structures, focus on local justifications, or lack scalability across diverse scenarios. They rarely capture holistic behavioral patterns or leverage broader common sense knowledge. These limitations motivate our textual-abstractive paradigm, where LLMs synthesize coherent global summaries of agent behavior.

\begin{figure}[th]
\centering
\includegraphics[width=\columnwidth]{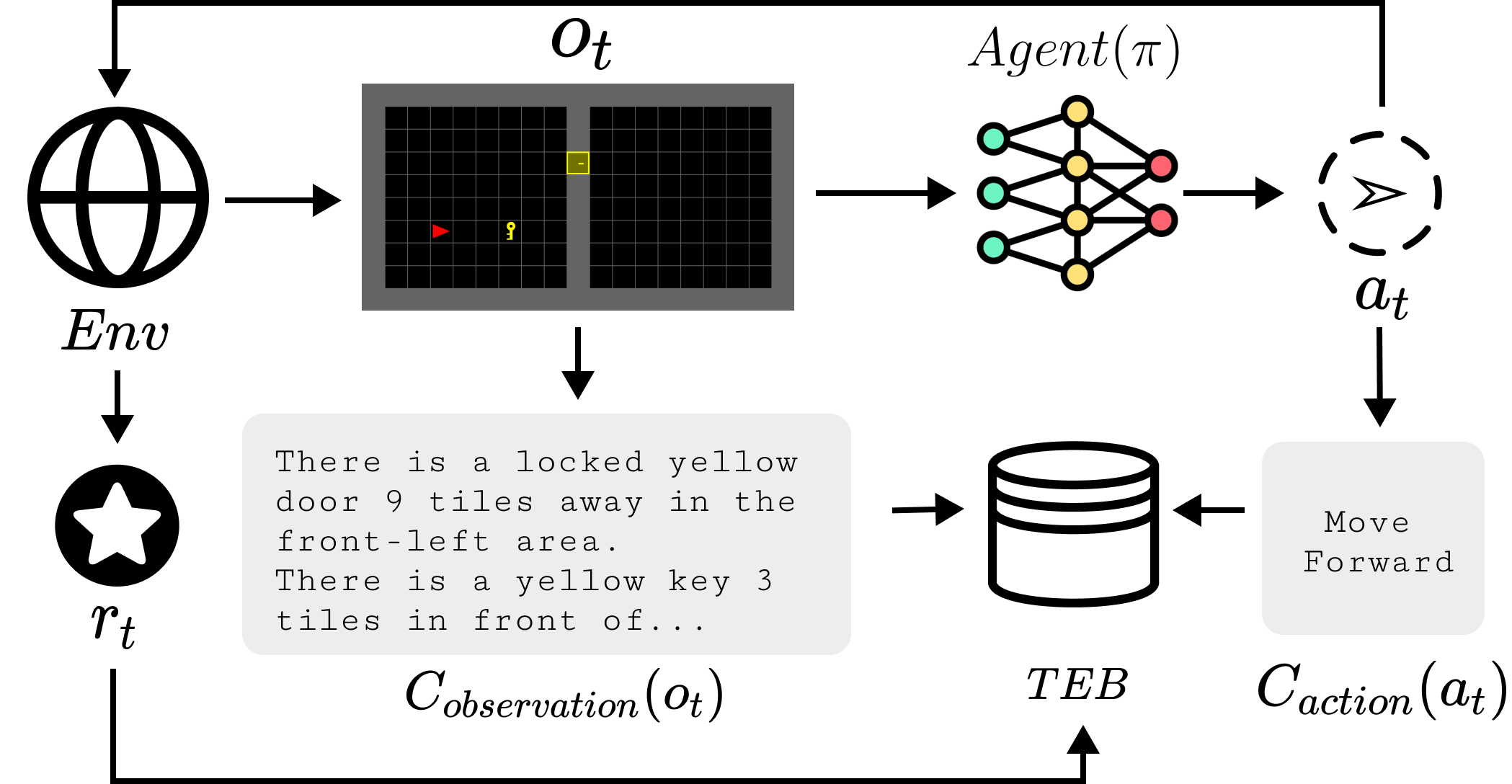}
\Description{Diagram of the data collection pipeline. The environment produces an observation grid showing objects like a yellow key and door. The observation is converted into a text description, then passed to an experience dataset. The agent’s policy network takes the observation and outputs an action, such as moving forward. The action and observation captions are stored in the dataset along with rewards.}
\caption{Collecting the textual experience buffer (Section~\ref{sec:experience}).}
\label{fig:SLLM methodology}
\end{figure}
\section{Problem Formulation}
In this section, we formalize the abstractive-summarization problem. Specifically, we describe the setting, define the summarization task, and introduce conceptual principles that guide both our methodology and our evaluation. 

\subsection{Setting}
We consider an RL environment modeled as a partially observable Markov decision process (POMDP):  
\[
\mathcal{M} = \langle \mathcal{S}, \mathcal{A}, \mathcal{O}, T, O, R, \gamma \rangle,
\]
where $\mathcal{S}$ is the state space, $\mathcal{A}$ the action space, $\mathcal{O}$ the observation space, 
$T(s' \mid s,a)$ the transition kernel, $O(o \mid s,a)$ the observation function, 
$R(s,a)$ the reward function, and $\gamma \in (0,1]$ the discount factor.  

An agent follows a stochastic policy
\[
\pi: \mathcal{O} \to \Delta(\mathcal{A}),
\]
which maps each observation $o \in \mathcal{O}$ to a distribution over actions.  
The interaction induces a distribution over trajectories
\[
\tau = (o_0, a_0, r_0, \dots, o_T), \quad \tau \sim \pi, T, O.
\]

For summarization, we assume access not only to isolated trajectories but to an \emph{experience buffer}
\[
\mathcal{B}_\pi = \{\tau_1, \dots, \tau_N\},
\]
which aggregates multiple episodes sampled from $\pi$. This buffer serves as the raw material based on which textual explanations are generated.  

\subsection{Policy Summarization Task}
We define a \emph{policy summarizer} as a mapping
\[
f: \mathcal{B}_\pi \to \mathcal{T},
\]
where $\mathcal{T}$ denotes the space of abstractive textual explanations.  

We require $f$ to approximate three key principles.  
\textbf{Expressiveness}: capture recurring behavioral patterns (e.g. ``the agent prioritizes unlocking doors before exploring rooms'') rather than isolated actions.  
\textbf{Scalability}: operate over long horizons and large buffers while maintaining concise summaries.  
\textbf{Faithfulness}: reflect the actual distribution of behaviors under $\pi$, avoiding hallucinated or spurious strategies.  

\paragraph{Conceptual Optimization View}
We frame summarization as a conceptual optimization problem:
\[
T^* = \arg\max_{T \in \mathcal{T}} \; \mathcal{U}(T \mid \mathcal{B}_\pi),
\]
where $\mathcal{U}$ is a utility function balancing:
\emph{coverage} (operationalizing expressiveness by accounting for recurring behaviors),
\emph{parsimony} (operationalizing scalability through concise abstraction), and
\emph{fidelity} (operationalizing faithfulness via alignment with empirical evidence in $\mathcal{B}_\pi$).

This formulation is not solved directly. Instead, it provides a guiding lens for both \emph{methodology} and \emph{evaluation}:  
SySLLM instantiates coverage through multi-trajectory aggregation and hierarchical summarization, parsimony through representative summary selection, and fidelity through expert alignment and user validation.  
In the following sections, we detail how these principles are operationalized in practice and evaluated empirically.

\begin{figure}[tbh]
\centering
  \includegraphics[width=\columnwidth]{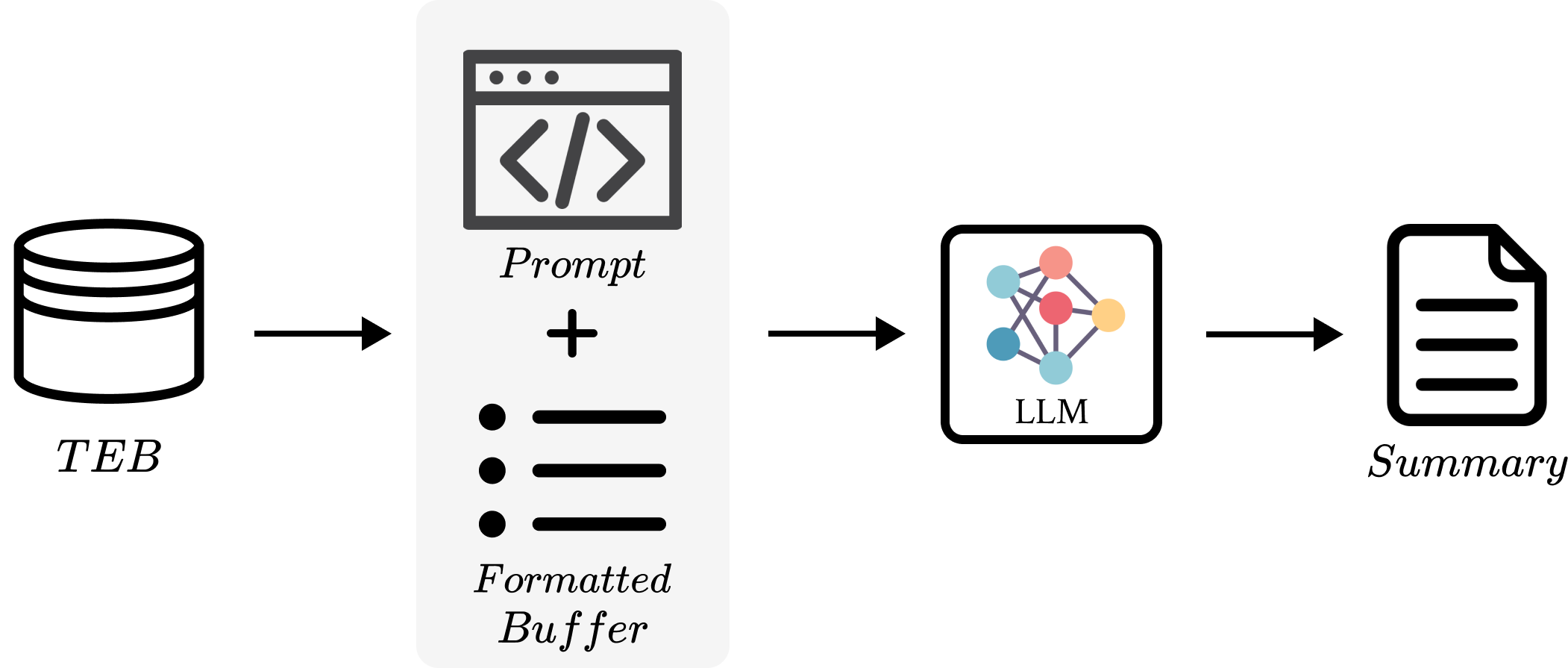}
  \Description{SySLLM generating policy summaries}
  \caption{Generating global policy summaries (Section~\ref{sec:summary}).}~\label{fig:SLLM methodology 2}
\end{figure}
\section{S\lowercase{y}SLLM Framework}
\label{sec:alg}

Our SySLLM (\textbf{Sy}nthesized \textbf{S}ummary using \textbf{LLM}s) framework formalizes policy summarization as a two-phase process: 
(i) \emph{experience collection and captioning}, which transforms trajectories into natural language traces stored in a \emph{Textual Experience Buffer (TEB)}, and (ii) \emph{abstractive summarization}, which synthesizes the TEB into a concise global description of the agent’s policy using a large language model (LLM). 
Algorithms~\ref{alg:algorithm}--\ref{alg:hier} specify the pipeline, while Figures~\ref{fig:SLLM methodology}--\ref{fig:SLLM methodology 2} illustrate its two phases.

\subsection{Captioners}
\label{sec:captioner}

The first phase converts raw trajectories into textualized experiences.  
For a trajectory $\tau = (o_1,a_1,r_1,\ldots,o_T)$, we define two captioning functions:
\[
C_{\text{obs}}: \mathcal{O} \to \Sigma^\star,
\quad
C_{\text{act}}: \mathcal{A} \to \Sigma^\star,
\]
where $\Sigma^\star$ denotes the set of natural language strings.  
At each step $t$, the pair $(o_t, a_t)$ is mapped to $(C_{\text{obs}}(o_t), C_{\text{act}}(a_t))$, yielding a textual description of the observation and the action taken.

\textit{Observation Captioner.}  
$C_{\text{obs}}$ produces structured descriptions of salient percepts (e.g. ``The agent is facing a locked door with a key to the left.'').  

\textit{Action Captioner.}  
$C_{\text{act}}$ verbalizes the agent’s action (e.g. ``move forward,'' ``pick up the key.'').  

This follows prior work on language rounding~\cite{jiang2019language,mirchandani2021ella}.  
In practice, captioners may be rule-based, vision-language based, or hybrid;  
the concrete instantiation is left as a domain-specific choice
and is discussed further in Section~\ref{sec:disc}.  

\subsection{Constructing the Textual Experience Buffer}
\label{sec:experience}

From the $N$ sampled episodes, we construct a \emph{Textual Experience Buffer (TEB)}:

\[
TEB_\pi = \langle e_1, e_2, \dots, e_L \rangle,
\quad
e_t = \langle \hat{o}_t, \hat{a}_t, r_t, epID \rangle,
\]
where $\hat{o}_t = C_{\text{obs}}(o_t)$ and $\hat{a}_t = C_{\text{act}}(a_t)$ denote natural language captions of observations and actions.  
Each tuple records captioned observation, captioned action, reward, and episode identifier, preserving temporal coherence between steps.  

Episodes are generated by sampling from the policy $\pi$:
\[
o_1 \sim \texttt{RESET}(ENV), \quad
a_t \sim \pi(o_t), \quad
(o_{t+1}, r_t, done) \sim \texttt{STEP}(a_t),
\]
repeating until $done = \texttt{True}$. At each step, both observation and action are passed through captioning functions $C_{\text{obs}}$ and $C_{\text{act}}$.  

The TEB is thus a textual analogue of a replay buffer: it aggregates multiple trajectories in a structured, language-based format that is directly consumable by LLMs, while maintaining the sequential structure necessary for policy-level reasoning. 
Further details on the fields stored in the TEB is provided in Appendix~\ref{apx:experience_dataset}.

\subsection{Abstractive Summarization}
\label{sec:summary}

Let $\mathcal{B}$ denote the space of textual experience buffers, where each
$TEB_\pi \in \mathcal{B}$ is a finite ordered sequence of captioned experience
tuples collected from executions of policy $\pi$.

The second phase maps the TEB to a global narrative $T \in \mathcal{T}$.  
We define the summarizer as
\[
f_{\theta} : \mathcal{B} \rightarrow \mathcal{T}, \qquad
T = f_{\theta}(TEB_\pi),
\]
where $\Sigma^\star$ denotes the space of finite token sequences and
$f_{\theta}$ is instantiated by an LLM conditioned on a structured prompt.


\emph{Prompt Construction.}  
The prompt follows a hierarchical structure, inspired by Chain-of-Thought reasoning~\cite{wei2022chain,kojima2022large}, 
which decomposes complex behavioral analysis into specific components.  
This design guides the LLM to progressively move from low-level traces in the TEB
to higher-level abstractions of the policy. The prompt consists of:
\begin{enumerate}[leftmargin=*]
    \item \textbf{General Instructions}: define the summarization task.  
    \item \textbf{Environment Context}: describe task objectives and constraints.
    \item \textbf{Textual Experience Buffer}: provide $TEB$ in structured form to preserve sequentiality.  
    \item \textbf{Output Specification}: constrain the output to a concise summary in natural language.  
\end{enumerate}
Prompt design details are provided in Appendix~\ref{apx:prompt_design}.

\emph{Scalability via Hierarchical Summarization.}  
A fundamental challenge in LLM-based summarization is the bounded context size: The TEB may exceed the maximum token budget $\kappa$ of the model.  
To address this, SySLLM employs a hierarchical procedure that recursively reduces
the buffer until it fits within the context window.  

Formally, define a summarization operator:
\[
\mathcal{S}_\theta : \mathcal{D} \to \Sigma^\star,
\]
where $\mathcal{D}$ is any subset of the TEB and $\Sigma^\star$ the space of textual summaries.  
If $|TEB| \leq \kappa$, we directly apply:
\[
T = \mathcal{S}_\theta(TEB).
\]
If $|TEB| > \kappa$, we partition the buffer into $M$ disjoint subsets 
$TEB = \{TEB^{(1)}, \dots, TEB^{(M)}\}$ such that each $|TEB^{(i)}| \leq \kappa$.  
For each subset we compute intermediate summaries:
\[
S_i = \mathcal{S}_\theta(TEB^{(i)}), \quad i=1,\dots,M,
\]
and aggregate them by applying $\mathcal{S}_\theta$ again:
\[
T = \mathcal{S}_\theta(\{S_1, \dots, S_M\}).
\]
This recursive divide-and-conquer scheme:
\[
T = \textsc{HierarchicalSummarize}(TEB, \kappa)
\]
ensures that SySLLM remains applicable to arbitrarily large buffers while
preserving coverage across all episodes and compressing details into intermediate summaries.  

\emph{Candidate Generation and Selection.}  
Once the input (original or hierarchical) fits within context, we query the LLM to generate $K$ candidate summaries $\{T_1, T_2, \dots, T_K\}$ via stochastic decoding.  
This captures variability in abstraction and phrasing.  
To select a robust final summary, each candidate is embedded into a semantic vector space using a pretrained embedding model, yielding $\phi(T_i) \in \mathbb{R}^d$.  
We compute the centroid:
\[
c = \tfrac{1}{K} \sum_{i=1}^K \phi(T_i),
\]
and measure distances:
\[
d_i = \lVert \phi(T_i) - c \rVert_2.
\]
The final summary $T^*$ is chosen as the \emph{median representative}, i.e.  
the candidate closest to the median-ranked distance from the centroid:
\[
T^* = \arg\min_{T_i} \; \big| \text{rank}(d_i) - \tfrac{K}{2} \big|.
\]

This selection scheme balances \emph{generality} (summaries near the centroid capture broad regularities) with \emph{specificity} (summaries farther away capture contextual details), yielding a consensus-style narrative.

\begin{algorithm}[tb]
\caption{\textsc{SySLLM Framework}}
\label{alg:algorithm}
\raggedright
\textbf{Input}: Environment $ENV$, trained policy $\pi$, captioners $C_{\text{obs}}$, $C_{\text{act}}$, base prompt $P$ \\
\textbf{Parameters}: number of episodes $N$, token budget $\kappa$, \#candidates $K$, embedding model $\phi(\cdot)$ \\
\textbf{Output}: Policy summary $T^* \in \mathcal{T}$

\begin{algorithmic}[1]
\STATE \textbf{Initialize} Textual Experience Buffer $TEB \gets \varnothing$ \label{alg:init_ teb}
\FOR{$i = 1$ \TO $N$} \label{alg:for_loop}
    \STATE $t \gets 1$; $o_t \gets ENV.\texttt{RESET}()$; $epReward \gets 0$; $done \gets \textbf{False}$ \label{alg:env_reset}
    \WHILE{$\neg done$} \label{alg:while_loop}
        \STATE $a_t \sim \pi(\cdot \mid o_t)$ \label{alg:sample_action}
        \STATE $TEB.\texttt{ADD}\big(\,C_{\text{obs}}(o_t),\, C_{\text{act}}(a_t),\, epReward,\, i\,\big)$ \label{alg:add_TEB}
        \STATE $(o_{t+1}, r_{t+1}, done) \gets ENV.\texttt{STEP}(a_t)$ \label{alg:step}
        \STATE $epReward \gets epReward + r_{t+1}$; $t \gets t + 1$ \label{alg:update}
        \STATE $o_t \gets o_{t+1}$
    \ENDWHILE
\ENDFOR \label{alg:end_phase_1}

\STATE $T \gets \textsc{HierarchicalSummarize}(TEB, \kappa, P, K, \phi)$ \label{alg:hier_call}
\STATE \textbf{return} $T$ \label{alg:return}
\end{algorithmic}
\end{algorithm}

\begin{algorithm}[tb]
\caption{\textsc{HierarchicalSummarize}}
\label{alg:hier}
\raggedright
\textbf{Input}: textual experience buffer subset $D$, token budget $\kappa$, prompt $P$, \#candidates $K$, embedding model $\phi$ \\
\textbf{Output}: summary $T \in \mathcal{T}$

\begin{algorithmic}[1]
\IF{$\texttt{TOKENS}(\texttt{FORMAT}(D)) \le \kappa$} \label{alg:fits}
    \STATE $X \gets P \;+\; \texttt{FORMAT}(D)$ \label{alg:format}
    \STATE $\{T_1,\ldots,T_K\} \gets \texttt{LLM.SAMPLE}(X, K)$ \label{alg:sampleK}
    \STATE $E_i \gets \phi(T_i) \in \mathbb{R}^d \quad \forall i \in \{1,\ldots,K\}$ \label{alg:embed}
    \STATE $c \gets \frac{1}{K}\sum_{i=1}^{K} E_i$ \label{alg:centroid}
    \STATE $d_i \gets \lVert E_i - c \rVert_2 \quad \forall i$; \; $j^\star \gets \arg\min_{j}\left|\,\texttt{rank}(d_j) - \frac{K}{2}\,\right|$ \label{alg:median}
    \STATE \textbf{return} $T_{j^\star}$ \label{alg:return_leaf}
\ELSE
    \STATE $\{D^{(1)},\ldots,D^{(M)}\} \gets \texttt{PARTITION}(D,\kappa)$ \hfill $\triangleright$ disjoint, each fits \label{alg:partition}
    \FOR{$m=1$ \TO $M$}
        \STATE $S_m \gets \textsc{HierarchicalSummarize}(D^{(m)}, \kappa, P, K, \phi)$ \label{alg:recurse}
    \ENDFOR
    \STATE \textbf{return} \textsc{HierarchicalSummarize}$\big(\{S_1,\ldots,S_M\}, \kappa, P, K, \phi\big)$ \label{alg:aggregate}
\ENDIF
\end{algorithmic}
\end{algorithm}
\section{Experimental Setup}
\label{sec:exp_setup}
We evaluate SySLLM across controlled reinforcement learning environments that capture both simple and complex agent behaviors. Specifically, we apply our framework to five environments from the MiniGrid suite~\cite{chevalier2021minimalistic} and the more challenging Crafter environment~\cite{hafner2021crafter}. These environments were selected to span a range of task structures, observation modalities, and policy complexities.
We use the \texttt{gpt-4-turbo} model with a temperature of 0.5~\cite{achiam2023gpt}, and the \texttt{text-embedding-3-small} model~\cite{openai2024embedding} as the embedding function.

\subsection{MiniGrid}
MiniGrid is a grid-world framework where agents perform goal-directed navigation and object-interaction tasks under partial observability. We instantiate seven agents across five environments, ensuring diversity in both policy performance and behavioral style.

\emph{Captioners.}
To construct the Textual Experience Buffer (TEB), we implement a rule-based captioning system that maps raw grid observations and discrete actions into structured natural-language descriptions. 
\textit{Observation Captioner} ($C_{\text{obs}}$): generates textual descriptions of visible elements, including object types (e.g., keys, doors, obstacles) and their spatial relationships relative to the agent.
\textit{Action Captioner} ($C_{\text{act}}$): translates the discrete action set (e.g., \texttt{turn\_left}, \texttt{move\_forward}) into natural-language strings.  
\emph{Agents and Training}
We trained three agents with qualitatively distinct policies in the \texttt{MiniGrid-Unlock} environment:  
\begin{itemize}[leftmargin=*]
    \item \textbf{Goal-directed agent}: wide $7 \times 7$ observation window, optimized to minimize the steps to unlock the door.  
    \item \textbf{Short-sighted agent}: restricted $3 \times 3$ observation window, leading to more myopic strategies.  
    \item \textbf{Random agent}: selects actions uniformly at random, providing a non-structured behavioral baseline.  
\end{itemize}
In addition, we trained one agent each in four further environments: \texttt{Dynamic Obstacles}, \texttt{Lava Gap}, \texttt{Red-Blue Doors}, and \texttt{Crossing}. All MiniGrid agents were trained using PPO~\cite{schulman2017proximal,stable-baselines3} for 1M timesteps per seed for three random seeds. Agent performance statistics are reported in Table~\ref{tab:agent_performance}. The complete hyperparameters are provided in the Appendix~\ref{apx:implementation_details}.

\emph{TEB Collection and Summarization.}
For each agent, we collect $50$ evaluation episodes to construct the TEB. Each buffer is formatted into the structured prompt and processed as described in Section~\ref{sec:summary}, where candidate summaries are generated via stochastic decoding and post-processed for abstractive summarization.
Prompt templates used for the MiniGrid suite domain are listed in Appendix~\ref{apx:sum_prompt}.

\subsection{Crafter}
Crafter is a 2D, partially observable world inspired by Minecraft, featuring procedurally generated maps, resource gathering, crafting, and an achievement tree that defines agent progress.

\begin{figure}[htb]
\centering
\includegraphics[width=\columnwidth]{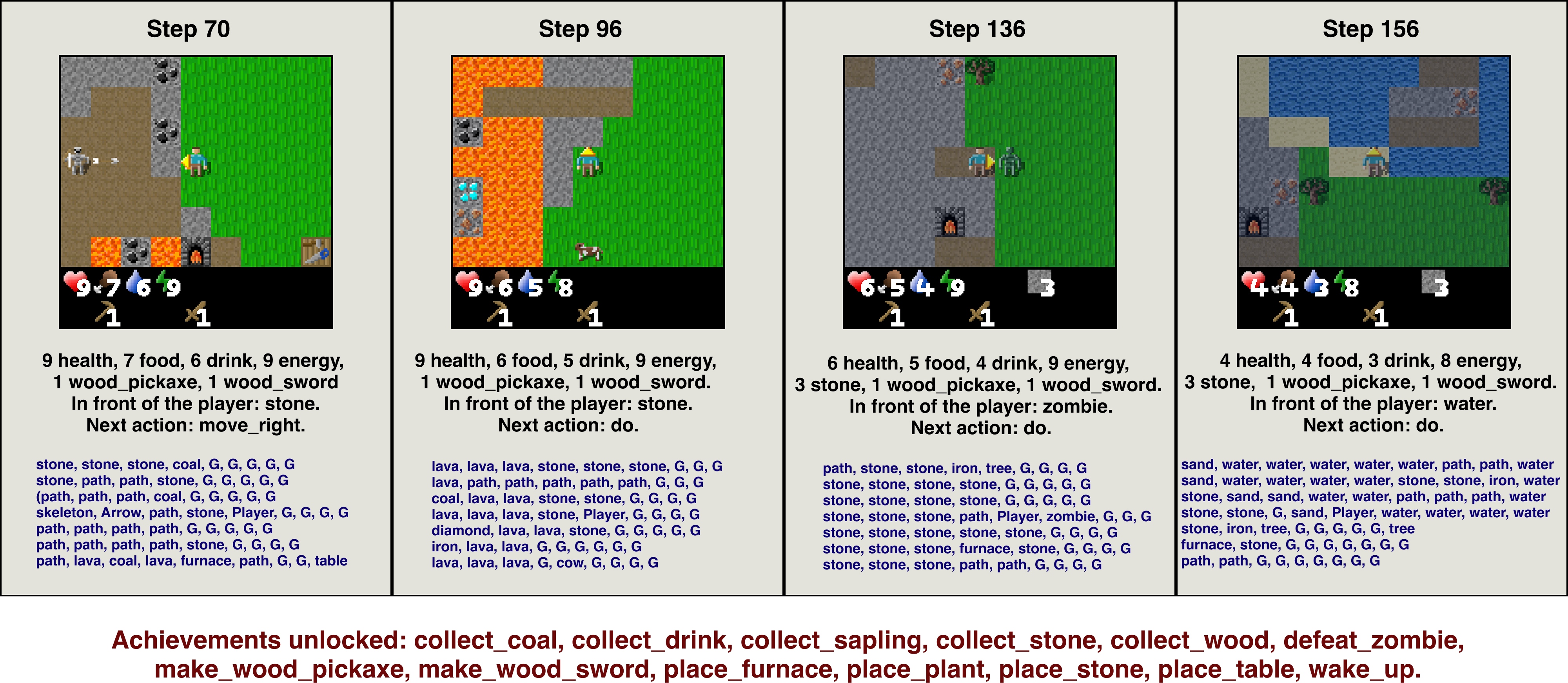}
\Description{Crafter captioning}
\caption{Four steps from a trajectory of the Resource-Collector agent in the Crafter environment, alongside their corresponding  captions generated using the observation and action captioners. For each step, the captions describe the agent's inventory status, the object currently in front of it, and the next action selected by the agent. A textual representation of the visible grid (highlighted in blue) is also included to reflect the agent's local perception. Additionally, all unique achievements unlocked by the agent throughout the trajectory are summarized in red.}
\label{fig:Crafter_figure}
\end{figure}

\emph{Captioners}
To adapt the captioning system to Crafter, we extend the MiniGrid captioner to encode:
\textit{Observation Captioner}: inventory contents, spatial relations to nearby resources and threats, current health and stamina, and unlocked achievements.
\textit{Action Captioner}: maps the Crafter action set into textual forms, e.g., \textit{``move\_right''}, \textit{``place\_table''}, \textit{``make\_wood\_sword''}.
Figure~\ref{fig:Crafter_figure} shows an example trajectory and its textualized representation.

\emph{Agents and Training}
We train two agents with distinct behavior:  
\begin{itemize}[leftmargin=*]
    \item \textbf{Resource-Collector agent}: trained with DreamerV3~\cite{hafner2023dreamerv3}, capable of sustained survival, resource collection, and the crafting of basic tools.  
    \item \textbf{Random agent}: uniformly samples from the action set, serving as a baseline with no structured policy.  
\end{itemize}

\emph{TEB Collection and Summarization.}
For each agent, we log $5$ evaluation episodes. Due to the long horizon of Crafter, the TEB for each episode can exceed the LLM's context length. In this case, we apply the hierarchical summarization strategy described in Section~\ref{sec:summary}. 
Prompt templates used for the Crafter domain are listed in Appendix~\ref{apx:crafter_prompts}.

\begin{table}[ht]
    \centering
    \resizebox{\columnwidth}{!}{
    \begin{tabular}{ccccc}
    \toprule
    \textbf{Env.} & \textbf{Agent} & \textbf{Mean Reward ± SD} & \textbf{Mean Length} & \textbf{Success / Achievements} \\
    \midrule
    \multirow{7}{*}{MiniGrid}
        & Unlock Goal-directed & 0.73 ± 0.21 & 20.25 & Success 0.93 \\
        & Unlock Short-sighted & 0.41 ± 0.27 & 44.43 & Success 0.77 \\
        & Unlock Random        & 0.00 ± 0.01 & 70.00 & Success 0.00 \\
        & Dynamic Obstacles    & 0.78 ± 0.06 & 17.20 & Success 1.00 \\
        & Lava Gap             & 0.90 ± 0.02 & 10.82 & Success 1.00 \\
        & Red Blue Doors       & 0.70 ± 0.26 & 17.06 & Success 0.88 \\
        & Crossing             & 0.67 ± 0.18 & 24.80 & Success 0.94 \\
    \midrule
    \multirow{2}{*}{Crafter}
        & Resource-Collector   & 10.43 ± 2.11 & 234.6 & 11.33 achievements \\
        & Random               & 1.39 ± 1.19  & 164.4 & 2.29 achievements \\
    \bottomrule
    \end{tabular}}
    \caption{Performance metrics across MiniGrid (500 episodes, 3 seeds) and Crafter (100 episodes, 3 seeds). For MiniGrid, performance is measured by success rate; for Crafter, by the number of unique achievements unlocked.}
    \label{tab:agent_performance}
\end{table}
\section{Illustrative Policy Summaries}
\label{sec:qualitative}

To illustrate the summaries produced by \sllm{}, we present structured case studies that connect textual motifs to quantitative evidence and then expand them into detailed narrative accounts. This combination shows that the framework produces explanations that are not only linguistically coherent but also grounded in the actual behavior of the agents. A full summary of the Unlock Goal-Directed agent is included in Appendix~\ref{apx:example_summary}.

\emph{Structured Case Studies.}
Table~\ref{tab:case_studies} reports representative agents in MiniGrid and Crafter. For each, we show: (i) the central claim extracted by SySLLM, and (ii) the quantitative metrics that support or refute this claim. The alignment between narrative motifs and behavioral statistics illustrates that SySLLM captures recurring strategies and limitations in a manner consistent with ground-truth agent performance. For example, the Unlock Goal-Directed agent's motif of ``turning toward the nearest key or door'' is supported by a $0.93$ success rate and shorter episode lengths, while the Random agent's ``lack of coherent strategy'' corresponds to maximal episode lengths and zero success rate. In Crafter, the Resource-Collector agent's emphasis on survival resources is matched by high reward and achievement counts, whereas the Random agent's incoherence is validated by minimal achievement progression.  

\begin{table*}[ht]
    \centering
    \resizebox{\linewidth}{!}{
    \begin{tabular}{cccc}
    \toprule
    \textbf{Env.} & \textbf{Agent} & \textbf{SySLLM Summary Claim} & \textbf{Quantitative Alignment} \\
    \midrule
    \multirow{4}{*}{MiniGrid} 
      & Unlock Goal-directed & Turns toward nearest key/door & Success $0.93$, mean length $20.25$ \\
      & Unlock Random & No clear strategy & Success $0.00$, max length episodes \\
      & Dynamic Obstacles & Avoids obstacles, occasional inefficiency & Success $1.00$, higher variance in length \\
      & Lava Gap & Jumps gap consistently & Success $1.00$, mean length $10.82$ \\
    \midrule
    \multirow{2}{*}{Crafter} 
      & Resource-Collector &  Focused, survival-oriented strategy; Unlocks early achievements & Reward $10.43$, $11.33$ unique achievements \\
      & Random & Sporadic, incoherent progression & Reward $1.39$, $2.29$ unique achievements \\
    \bottomrule
    \end{tabular}}
    \caption{Structured case studies of SySLLM summaries across environments. Extracted motifs are validated against quantitative metrics, showing alignment between narrative claims and behavioral statistics.}
    \label{tab:case_studies}
\end{table*}

\emph{Expanded Narrative Insights.}
Beyond single-sentence claims, SySLLM produces multi-faceted descriptions of agent behavior. Figure~\ref{fig:insights} presents detailed motifs across environments, highlighting how the framework abstracts local decisions into global patterns. These narratives capture not only the strengths of agents (e.g. consistent lava avoidance, structured door-order strategies) but also nuanced inefficiencies (e.g., unnecessary turns in dense obstacle scenarios, repetitive failed crafting attempts in Crafter). Such fine-grained insights demonstrate SySLLM's ability to diagnose effective heuristics and characteristic failure modes.  

\begin{figure*}[t]
\scriptsize
\begin{multicols}{3}
\textbf{Unlock – Goal-directed}
\begin{enumerate}[leftmargin=*]
\item \textit{The agent effectively identifies keys and adjusts its path based on their relative position, shifting focus to unlocking the door.}
\item \textit{It consistently turns towards the nearest key or door, minimizing distance, which remains consistent across episodes.}
\item \textit{The agent completes episodes efficiently, averaging 15--25 steps with near-maximal cumulative rewards.}
\end{enumerate}

\textbf{Unlock – Short-sighted}
\begin{enumerate}[leftmargin=*]
\item \textit{The agent follows a right-wall method, moving forward until encountering an obstacle before turning.}
\item \textit{It identifies keys and doors efficiently, maneuvering toward and using them correctly.}
\item \textit{Decisions are heavily influenced by its immediate field of vision, reacting only to nearby objects.}
\end{enumerate}

\textbf{Unlock – Random}
\begin{enumerate}[leftmargin=*]
\item \textit{The agent exhibits unstructured behavior, often repeating unnecessary actions.}
\item \textit{It frequently toggles doors multiple times or picks up and drops keys without using them effectively.}
\end{enumerate}

\textbf{Lava Gap}
\begin{enumerate}[leftmargin=*]
\item \textit{The agent consistently avoids lava, demonstrating awareness of environmental hazards.}
\item \textit{Upon encountering an obstacle, it either turns or moves in the opposite direction.}
\end{enumerate}

\textbf{Red--Blue}
\begin{enumerate}[leftmargin=*]
\item \textit{The agent prioritizes opening the red door before the blue door, optimizing reward accumulation.}
\item \textit{It successfully interacts with doors in a structured sequence, adhering to task constraints.}
\end{enumerate}

\textbf{Crossing}
\begin{enumerate}[leftmargin=*]
\item \textit{The agent moves towards the green goal once it enters its field of vision, adjusting its path accordingly.}
\item \textit{It avoids collisions with walls through timely directional changes.}
\end{enumerate}

\textbf{Dynamic Obstacles}
\begin{enumerate}[leftmargin=*]
\item \textit{The agent effectively avoids moving obstacles (blue balls) by adjusting its movement.}
\item \textit{It identifies objects in its field of vision and makes informed navigation decisions.}
\item \textit{In dense obstacle scenarios, occasional inefficiencies or unnecessary turns are observed.}
\end{enumerate}


\textbf{Crafter – Resource-collector}
\begin{enumerate}[leftmargin=*]
\item \textit{Strong Focus on Resource Collection: The agent consistently prioritizes gathering essential resources such as wood, stone, and drink, which are foundational for crafting tools and maintaining basic survival metrics.}
\item \textit{Effective Basic Tool Crafting: Regular crafting of basic tools like wood pickaxes and swords enables the agent to enhance resource collection and engage in occasional combat.}
\item \textit{Achievement Unlocking: The agent reliably unlocks achievements related to resource collection and basic tool crafting but struggles with more advanced achievements, highlighting a potential area for improvement.}
\item \textit{Moderate Combat Engagement: The agent occasionally engages with zombies, using crafted tools for defense, showing moderate adaptability to threats but limited combat readiness overall.}
\item \textit{Predictable Behavior: Episodes are characterized by high consistency in resource collection and basic tool crafting actions, with low variance across episodes.}
\item \textit{Inconsistent Health and Exploration Management: While the agent effectively manages drink levels, it shows less consistency in food and health management and sacrifices exploration efficiency for achievement unlocking.}
\end{enumerate}


\textbf{Crafter – Random}
\begin{enumerate}[leftmargin=*]
\item \textit{Ineffective Crafting and Resource Management: The agent frequently attempts crafting without the necessary resources or understanding of prerequisites, leading to repeated failures and minimal progress in achieving complex objectives.}
\item \textit{Poor Survival Strategy: The agent consistently demonstrates ineffective survival behavior, including health depletion and poor management of food and drink levels, which hampers its ability to sustain itself in the game environment.}
\item \textit{Limited Achievement Progression: While the agent reliably unlocks basic achievements like wake\_up and collect\_sapling, it struggles to achieve more complex milestones that require crafting, resource management, or combat engagement.}
\item \textit{Repetitive and Ineffective Actions: Episodes are marked by high frequencies of `noop` actions and repetitive failed attempts at crafting, reflecting a lack of strategic adaptation and learning from past failures.}
\item \textit{Lack of Combat Engagement: The agent shows minimal engagement with combat mechanics and fails to defend effectively against threats such as zombies and skeletons.}
\item \textit{Predictable Behavior: Across episodes, the agent exhibits consistent, ineffective patterns of action, suggesting significant limitations in its decision-making processes and adaptability.}
\end{enumerate}

\end{multicols}
\Description{Insights from agents' SySLLM summaries in the MiniGrid environments.}
\caption{Insights from agents' SySLLM summaries in the MiniGrid environments.}
\label{fig:insights}
\end{figure*}

Together, 
these examples demonstrate that abstractive textual summaries can function as faithful and interpretable accounts of reinforcement learning policies.

\section{Expert Evaluation of Summaries}
\label{sec:expert_eval}

We complement the quantitative performance analysis with an expert-based evaluation of \sllm{} summaries. The goal is to assess how well the generated summaries capture the behavioral motifs observed by human experts and to quantify correctness while identifying potential hallucinations. 

\emph{Evaluation Protocol.}
We recruited six graduate students with research experience in the training and evaluation of RL agents. The experts were divided into two groups: Experts 1--3 annotated the MiniGrid-Unlock (goal-directed and short-sighted) agents, while Experts 4--6 annotated the remaining MiniGrid agents. In addition, Experts 1, 2, and 6 annotated the Crafter agents. Each expert was shown a 120-second video per agent, depicting representative trajectories. Based on these trajectories, the experts were instructed to produce textual summaries using the same \sllm{} prompting guidelines (see Appendix~\ref{apx:experts_instructions}). This alignment ensures comparability between expert- and model-generated summaries. 

\emph{Scoring Framework.}
To compare \sllm{} summaries $S_{\text{LLM}}$ with expert summaries $\{S_{\text{exp}}^j\}_{j=1}^m$, we decompose both into sets of atomic key points, denoted $\mathcal{K}_{\text{LLM}}$ and $\mathcal{K}_{\text{exp}}^j$, respectively. Semantic equivalence is assessed at the level of atomic propositions, where each key point expresses a single behavioral claim. Following standard practice in evaluating human-interpretable explanations \citep{hoffman2018metrics}, propositions are manually extracted and matched without embedding-based or heuristic similarity measures. Each \sllm{} key point is labeled as a full match, partial match, or unsupported relative to the expert set. This procedure yields a transparent and reproducible matching process, with substantial inter-annotator agreement (AC1 = 0.72). The full annotation protocol is provided in the appendix.

Each pairwise comparison is scored as:
\[
\text{match}(k_{\text{LLM}}, k_{\text{exp}}) =
\begin{cases}
1 & \text{if semantically equivalent}, \\
0.5 & \text{if partially overlapping}, \\
0 & \text{otherwise}.
\end{cases}
\]

\emph{Recall.}
Recall measures the extent to which the model summary covers expert-identified key points:
\[
\text{Recall}
= \frac{1}{m} \sum_{j=1}^m
\frac{
\sum_{k \in \mathcal{K}_{\text{exp}}^j}
\max_{k' \in \mathcal{K}_{\text{LLM}}}
\text{match}(k, k')
}{
\lvert \mathcal{K}_{\text{exp}}^j \rvert
}.
\]

\emph{Precision.}
Precision measures the correctness of \sllm{} key points relative to expert judgments. For each expert $j$, annotators are shown the set difference $\mathcal{K}_{\text{LLM}} \setminus \mathcal{K}_{\text{exp}}^j$ and asked to label each key point as Matched, Partially Matched, or Not Matched. Precision is defined as:
\[
\text{Precision}
= \frac{
\sum_{k \in \mathcal{K}_{\text{LLM}}}
\max_{j}
\text{match}(k, \mathcal{K}_{\text{exp}}^j)
}{
\lvert \mathcal{K}_{\text{LLM}} \rvert
}.
\]

This formulation captures both coverage (recall) and correctness (precision), while enabling explicit identification of hallucinated content through unmatched model key points.

\emph{Results.}
Table~\ref{table:recall_precision_summary} reports per-agent recall and precision. Recall scores range from 0.687 (Unlock goal-directed) to 0.914 (Crossing), with a mean of 0.840, demonstrating substantial overlap with expert-identified points. Precision scores range from 0.769 (Dynamic Obstacles) to 0.864 (Short-Sighted), with a mean of 0.839, indicating minimal hallucination. For example, in the Crossing environment, the point \textit{``The agent frequently checks for walls in its path and adjacent tiles''} was rejected by all experts, illustrating a rare hallucination. 

\emph{Inter-Annotator Agreement.}
To ensure the reliability of expert annotations, we calculated both raw agreement (percentage of identical match scores across experts) and Gwet's AC1 coefficient, which is robust to class imbalance in categorical judgments. Across all agents, the mean raw agreement was 70\%, while the mean AC1 reached 0.72, indicating substantial inter-rater reliability. This ensures that the observed recall/precision metrics are not artifacts of inconsistent annotations.

Taken together, these results show that \sllm{} summaries exhibit both high coverage and correctness relative to expert annotations, with recall and precision consistently above 0.8.

\begin{table}[t]
\centering
\resizebox{\columnwidth}{!}{
\begin{tabular}{c|ccc|c}
\toprule
\textbf{Agent} & \textbf{Expert} & \textbf{Recall} & \textbf{Precision} & \textbf{Mean} \\
\midrule
\multirow{3}{*}{Unlock Goal-Directed} 
 & E1 & 0.500 & 0.864 & \multirow{3}{*}{R = \textbf{0.687}, P = \textbf{0.864}} \\
 & E2 & 0.643 & 0.864 & \\
 & E3 & 0.917 & 0.864 & \\
\cmidrule(lr){1-5}
\multirow{3}{*}{Unlock Short-Sighted} 
 & E1 & 0.800 & 0.846 & \multirow{3}{*}{R = \textbf{0.878}, P = \textbf{0.839}} \\
 & E2 & 0.833 & 0.807 & \\
 & E3 & 1.000 & 0.923 & \\
\cmidrule(lr){1-5}
\multirow{3}{*}{Dynamic Obstacles} 
 & E4 & 0.583 & 0.692 & \multirow{3}{*}{R = \textbf{0.739}, P = \textbf{0.769}} \\
 & E5 & 0.833 & 0.692 & \\
 & E6 & 0.800 & 0.923 & \\
\cmidrule(lr){1-5}
\multirow{3}{*}{Lava Gap} 
 & E4 & 0.667 & 0.769 & \multirow{3}{*}{R = \textbf{0.794}, P = \textbf{0.811}} \\
 & E5 & 0.786 & 0.846 & \\
 & E6 & 0.929 & 0.818 & \\
\cmidrule(lr){1-5}
\multirow{3}{*}{Red-Blue Doors} 
 & E4 & 0.857 & 0.767 & \multirow{3}{*}{R = \textbf{0.871}, P = \textbf{0.834}} \\
 & E5 & 0.857 & 0.867 & \\
 & E6 & 0.900 & 0.867 & \\
\cmidrule(lr){1-5}
\multirow{3}{*}{Crossing} 
 & E4 & 0.750 & 0.731 & \multirow{3}{*}{R = \textbf{0.914}, P = \textbf{0.795}} \\
 & E5 & 0.917 & 0.808 & \\
 & E6 & 1.000 & 0.846 & \\
\cmidrule(lr){1-5}
\multirow{3}{*}{Crafter Resource-Collector} 
 & E7 & 0.938 & 0.893 & \multirow{3}{*}{R = \textbf{0.931}, P = \textbf{0.871}} \\
 & E8 & 0.938 & 0.857 & \\
 & E9 & 0.917 & 0.864 & \\
\cmidrule(lr){1-5}
\multirow{3}{*}{Crafter Random} 
 & E7 & 0.929 & 0.917 & \multirow{3}{*}{R = \textbf{0.902}, P = \textbf{0.929}} \\
 & E8 & 1.000 & 0.958 & \\
 & E9 & 0.778 & 0.857 & \\
\bottomrule
\end{tabular}}
\caption{Per-expert recall and precision scores for \sllm{} summaries, with aggregated per-agent means. $R$ = Recall, $P$ = Precision. Overall averages across all agents: $R = 0.840$, $P = 0.839$.}
\label{table:recall_precision_summary}
\end{table}

\section{User Study}\label{sec:user study}
We conducted a controlled user study to evaluate the usefulness of \sllm{} summaries compared to HIGHLIGHTS-DIV (\hili{})~\cite{amir2018highlights}, a standard demonstration-based benchmark in XRL which selects a set of high importance and diverse execution trajectories. The study assessed both subjective preferences and objective task performance. We focus on three qualitatively distinct agents from the MiniGrid Unlock environment (goal-directed, short-sighted, and random), ensuring the diversity of policies from structured strategies to noisy behaviors. For completeness, we provide a short description of \hili{} in Appendix~\ref{apx:highlight}.  

\emph{Experimental Design.}  
The study used a mixed design with two tasks. Task~1 (Preferences) followed a within-subject setup: each participant evaluated both modalities (\sllm{} and \hili{}), with order counterbalanced to mitigate ordering effects. Task~2 (Identification) followed a between-subject setup: Participants viewed only one modality, aligned with their Task~1 order. This produced four experimental conditions that varied by summary modality and agent type, as shown in Table~\ref{tab:experimental_conditions}. For \hili{}, highlight videos were generated using 300 traces, a context length of 5, and 20 highlights, while \sllm{} summaries were generated from 50 captioned episodes (see Section~\ref{sec:exp_setup}).

\begin{table}[ht!]
    \centering
    \resizebox{\columnwidth}{!}{
    \begin{tabular}{|c|c|c|c|}
        \hline
        \textbf{Condition} & \textbf{Task 1 Sequence} & \textbf{Task 1 Agent Type} & \textbf{Task 2 Summary Type} \\
        1 & \sllm{} $\rightarrow$ \hili{} & Goal-directed Agent & \sllm{} \\
        2 & \sllm{} $\rightarrow$ \hili{} & Short-sighted Agent & \sllm{} \\
        3 & \hili{} $\rightarrow$ \sllm{} & Goal-directed Agent & \hili{} \\
        4 & \hili{} $\rightarrow$ \sllm{} & Short-sighted Agent & \hili{} \\
        \hline
    \end{tabular}
    }
    \caption{Experimental conditions.}
    \label{tab:experimental_conditions}
\end{table}

\emph{Procedure.}  
The participants first completed a tutorial on MiniGrid Unlock rules, followed by a comprehension quiz. In Task~1, they watched a 120 seconds video of an agent's behavior and then rated a summary (\sllm{} or \hili{}) on eight explanation quality metrics (7-point Likert), adapted from Hoffman et al.~\cite{hoffman2018metrics}. After evaluating both modalities, they provided direct preference judgments: which summary better reflected the agent's policy and by what margin. In Task~2, participants were shown a summary (textual or visual) and asked to match it to one of three short (20s) videos: the correct agent plus two distractors. Each participant completed three trials ($Q1$: goal-directed, $Q2$: random, $Q3$: short-sighted). For each, they indicated their choice, rated confidence, and provided a justification.  

\emph{Participants.}  
We recruited 200 participants from Prolific (native English speakers from the US, UK, Canada, and Australia). Compensation was £3.75 base plus a £1 bonus for correct completion of Task~2. After exclusions for failed attention checks and implausibly short completion times (below 300 seconds), 192 participants remained (94 female, $M_{age}=36.4$, $SD=12.1$).  

\begin{figure}[ht]
    \centering
    \includegraphics[width=\columnwidth]{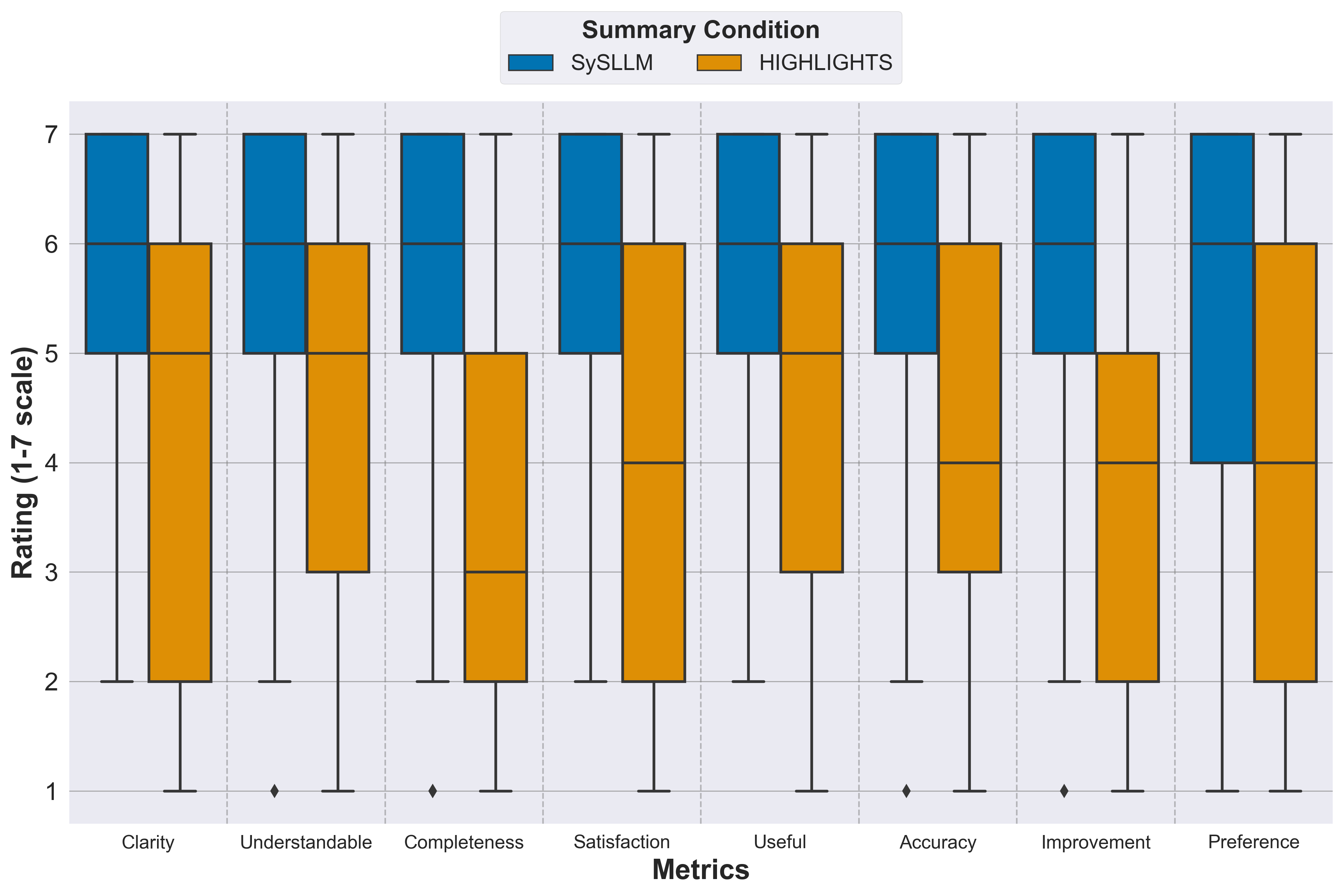}
    \Description{Participant scores for Task 1.}
    \caption{Participant ratings for Task 1 on a 1--7 Likert scale. \sllm{} ratings are significantly higher than \hili{} ratings across all metrics.}
    \label{fig:plot_task1}
\end{figure}

\begin{figure}[ht]
    \centering
    \includegraphics[width=\columnwidth]{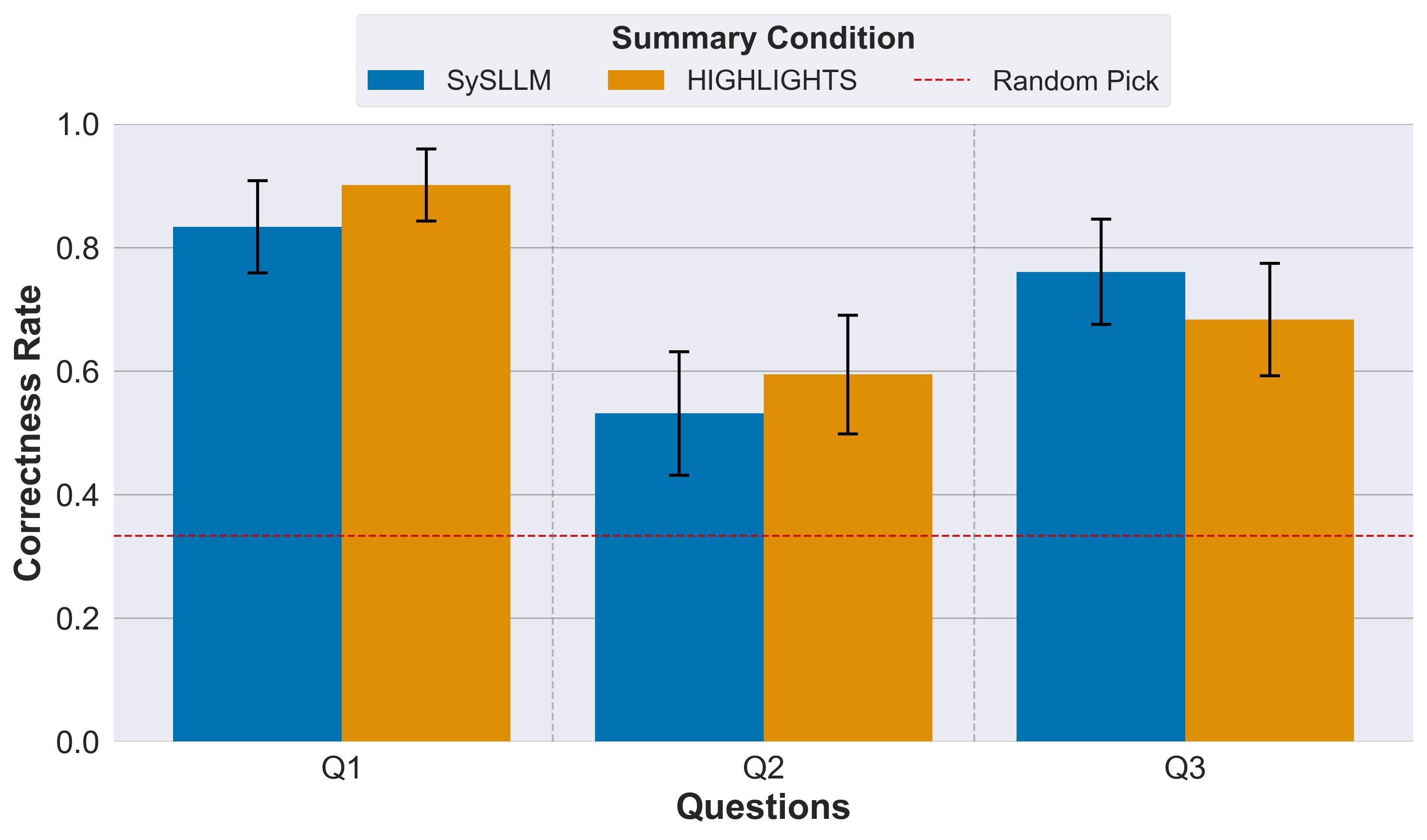} 
    \Description{Correctness rate in Task 2.}
    \caption{Correctness rate in Task 2. Error bars indicate 95\% confidence intervals. Both \sllm{} and \hili{} outperform the random guess baseline. The differences in correctness rate between \sllm{} and \hili{} is not statistically significant.}
    \label{fig:plot_task2}
\end{figure}

\emph{Results.}  
In Task~1, \sllm{} consistently outperformed \hili{} across all metrics (Fig.~\ref{fig:plot_task1}). Paired $t$-tests confirmed the difference as highly significant ($T=13.99$, $p < 10^{-33}$). Direct preference questions reinforced this: 75.5\% of participants favored \sllm{}, and the comparative Likert rating averaged $M=5.97$, $SD=1.44$ (neutral baseline = 4). Qualitative feedback highlighted that \sllm{} explained \textbf{why} agents acted as they did, while \hili{} required subjective inference. For example: \textit{``There are instances in the video where the agent seems to turn random corners. The summary explains why.''} 
In Task~2, correctness rates for both modalities exceeded random-guess baselines (Fig.~\ref{fig:plot_task2}). Chi-Square tests found no significant differences between \sllm{} and \hili{} across Q1–Q3. However, confidence scores revealed a significant effect in Q3, where participants in the \sllm{} condition reported higher confidence ($t=3.42$, $p=0.0008$).  

Overall, participants rated \sllm{} summaries as significantly clearer and more informative than the highlight videos. Although both modalities supported correct agent identification, textual summaries provided stronger interpretive cues, particularly reflected in higher confidence for certain agents. These results suggest that abstractive, language-based policy summaries enhance subjective interpretability while maintaining competitive performance in behavior recognition.

\section{Discussion and Future Work} \label{sec:disc}

We introduced \textsc{SySLLM}, a framework for \textit{abstractive textual} explanation of reinforcement learning policies. By converting state and action trajectories into structured language and leveraging large language models to generate policy summaries, \textsc{SySLLM} abstracts low-level decision traces into high-level behavioral patterns. Empirical results show close alignment with expert interpretations and a clear user preference over visual demonstrations, supporting language-based abstraction for interpreting complex policies.

Several limitations remain. \textsc{SySLLM} currently relies on domain-specific captioning functions to translate trajectories into text. While feasible in simulated environments, extending to high-dimensional or partially observed domains will require perceptual grounding via visual-language pipelines or pretrained vision-language models capable of zero-shot scene understanding.

Future work extends beyond static summarization. One direction is \textit{interactive policy querying}, where users ask natural language questions about agent behavior, enabling counterfactual and rationale-based explanations. Another is \textit{comparative summarization}, contrasting policies across training stages, reward functions, or architectures. Finally, integrating textual and visual modalities and advancing toward \textit{open-domain policy summarization} may enable benchmark-agnostic systems that characterize agent behavior at scale.

\begin{acks}
Funded by the European Union (ERC, Convey, 101078158) and the Israel Science Foundation (ISF) under Grant No. 3079/24. 
Views and opinions expressed are however those of the author(s) only and do not necessarily reflect those of the European Union or the European Research Council Executive Agency. Neither the European Union nor the granting authority can be held responsible for them.
\end{acks}



\bibliographystyle{ACM-Reference-Format} 
\bibliography{bibs/bibs}


\clearpage
\appendix
\section{HIGHLIGHTS}\label{apx:highlight}
\paragraph{``Highlights'' Policy Summaries} 
Our user study uses ``Highlights'' policy summaries~\cite{amir2018highlights} as a baseline. For completeness, we describe the algorithm here.
The HIGHLIGHTS algorithm generates an online summary of an agent’s behavior from simulations, using state importance to decide which states to include. A state is considered important if taking a wrong action there significantly decreases future rewards, as determined by the agent’s Q-values. Formally, state importance \(I(s)\) is defined as:
\begin{equation*}
\label{eq:importance}
I(s) = \max\limits_{a}Q^{\pi}_{(s,a)} - \min\limits_{a}Q^{\pi}_{(s,a)}.
\end{equation*}

HIGHLIGHTS captures trajectories with the most important states encountered in simulations. At each step, it evaluates state importance and adds the state to the summary if its importance exceeds the current minimum in the summary, replacing the least important state. For each state added, it also includes a trajectory of neighboring states and actions.

To address redundancy in similar important scenarios, the HIGHLIGHTS-DIV algorithm extends HIGHLIGHTS by incorporating diversity. HIGHLIGHTS-DIV evaluates a state \(s\) by identifying the most similar state \(s'\) in the summary. It compares \(I(s)\) to \(I(s')\) instead of the minimum importance value. If \(I(s)\) is greater, the trajectory including \(s'\) is replaced with the current trajectory. This approach maintains less important but diverse states, enhancing the information conveyed to users.

\section{Information Stored in Experience dataset}\label{apx:experience_dataset}
\begin{table}[H]
    \centering
    \begin{tabular}{|p{0.3\columnwidth}|p{0.60\columnwidth}|}
        \hline
        \textbf{Information} & \textbf{Description} \\
        \hline
        Episode Number & The number of the episode from which the data was collected. \\
        \hline
        Step Number & The specific step within the episode. \\
        \hline
        Captioned Observation & The observation converted into natural language. \\
        \hline
        Captioned Action & The action converted into natural language. \\
        \hline
        Cumulative Reward & The total reward accumulated by the agent up to that step. \\
        \hline
    \end{tabular}
    \caption{Description of the data stored in the experience dataset.}
\end{table}

\section{Summarization Prompt:}\label{apx:sum_prompt}
\begin{quote}
\ttfamily
[General Instructions]\\
Generate a focused summary of the RL agent’s policy based on the provided episodes data. Highlight key behaviors, decision-making processes, and patterns specific to this agent. Tailor the summary to reflect unique strategies and actions observed.\\
Focus on:\\
• Recurring patterns and behaviors specific to this agent’s policy.\\
• Detailed analysis of decision-making processes and responses to different stimuli.\\
• Efficiency in identifying and interacting with relevant objects (e.g., keys, doors).\\
• Methods used to solve tasks and handle obstacles.\\
• Comparison of agent's performance across different episodes.\\
• Quantitative metrics (e.g., number of steps, success rates) to evaluate efficiency.\\
• Analysis of navigation strategies and adaptations to the environment.\\
• Provide insights on the variability and randomness of the agent’s actions and decisions. Look at the distribution of the agent's actions during the episodes.\\
• Describe any notable fluctuations or inconsistencies in behavior across episodes.\\
• Figure out what is the agent's field of vision and its influence on its decisions and actions, considering it may not see the entire grid but only a part of it.\\

[Environment Description]\\
• Goal: Maximize cumulative reward (close to 1) by opening the door.\\
• The environment is static, with only the agent moving within it. Walls are fixed and form the boundaries of the grid (there are no further obstacles).\\
• Possible agent's actions: move forward, turn left, turn right, pick up, toggle, drop.\\
• Max episode length: 70 steps (failure to open the door if exceeded).\\

[Interpretation Instructions]\\
• If the next action is blank, the episode ended successfully. Use the cumulative reward to analyze performance.\\
• Follow the agent’s received observations to understand its behavior and decision-making process.\\
• Performance metrics: Number of steps to open the door, cumulative reward.\\
• Descriptions are from the perspective of the agent, with the number of tiles indicating the distance.\\

[Output Instructions]\\
Provide a concise summary focused on specific behaviors and actions of the agent. Formatted for easy understanding by non-technical individuals. Avoid general or trivial statements; focus on specific insights from the agent’s behavior. Include quantitative metrics and comparisons to provide a clear understanding of performance.\\

[Formatted Experience Dataset]\\
......................................................
\end{quote}

\section{Unlock Goal-Directed Agent Full Summary}\label{apx:example_summary}
\begin{quote}
This is an example of a summary produced by the SySLLM algorithm, illustrated here using the Unlock goal-directed agent:

\texttt{The analysis of the RL agent's behavior across 70 episodes reveals a consistent pattern in its decision-making and interaction with the environment. The agent's primary goal is to locate and use a key to open a corresponding colored door within a grid environment. Here are the key observations and insights from the agent's performance:}

\textbf{1. Navigation and Object Interaction:}  
\texttt{- The agent effectively identifies and navigates towards keys, demonstrating a capability to adjust its path based on the key's relative position. Once the key is picked up, the agent shifts its focus to finding and reaching the corresponding door.}  
\texttt{- The agent uses a combination of 'move forward', 'turn left', and 'turn right' actions to navigate towards objectives, showing proficiency in altering its path dynamically based on the environment's layout and the objects' positions.}  

\textbf{2. Efficiency Metrics:}  
\texttt{- The agent's efficiency varies across episodes, with some episodes achieving the goal in fewer steps and higher cumulative rewards.}  
\texttt{- The average steps taken to complete an episode when successful are approximately 15-25 steps, with cumulative rewards often nearing the maximum, indicating high efficiency in several scenarios.}  

\textbf{3. Decision-Making Patterns:}  
\texttt{- The agent demonstrates a pattern of turning towards the nearest key or door once identified, suggesting a straightforward heuristic of minimizing distance to the target. This behavior is consistent across episodes, indicating a robust underlying policy for basic navigation and objective fulfillment.}  
\texttt{- In scenarios where the agent is positioned with immediate obstacles or walls, it shows a tendency to turn and reorient itself towards open paths, reflecting an ability to avoid simple traps and dead ends.}  

\textbf{4. Handling of Obstacles:}  
\texttt{- The agent effectively navigates around walls and adjusts its path to reach its targets. There is no evidence of the agent being stuck or failing to navigate around corners, which suggests good spatial awareness and pathfinding within the grid constraints.}  

\textbf{5. Variability and Randomness:}  
\texttt{- The agent's actions show some variability, particularly in scenarios where multiple paths can lead to the objective. This variability might indicate a level of exploration or randomness in the policy, allowing the agent to discover and adapt to new optimal paths.}  

\textbf{6. Performance Consistency:}  
\texttt{- Across different episodes, the agent's performance shows consistency in terms of strategy but varies in execution efficiency, likely due to the differing initial conditions and environmental layouts. The agent consistently attempts to pick up the nearest key and unlock the door, aligning with its primary objective.}  

\textbf{7. Field of Vision Impact:}  
\texttt{- The agent's decisions are heavily influenced by its immediate field of vision, as it tends to react to objects as they become visible. This suggests a limited perception range, requiring the agent to physically move closer to distant objects to detect and interact with them.}  

\texttt{In summary, the RL agent is proficient in navigating grid environments to achieve specified goals of picking keys and opening doors. It demonstrates effective use of basic navigational actions, adapts well to the visible environment, and shows a consistent approach to achieving its objectives, albeit with some variability in efficiency possibly due to explorative behaviors or initial positioning. The agent's performance could potentially be enhanced by expanding its field of vision or optimizing its pathfinding algorithms to increase consistency in achieving higher rewards in fewer steps.}
\end{quote}

\section{Crafter Summarization Prompts} \label{apx:crafter_prompts}
\begin{ttfamily}
\noindent You are analyzing multiple episode summaries of an agent's behavior in the Crafter environment. Your goal is to identify consistent behavioral patterns across episodes and develop an objective behavioral profile.

\noindent [Cross-Episode Analysis Framework]

\noindent Systematically evaluate these key dimensions:

\noindent Behavioral Consistency

- Identify actions and strategies that appear consistently across episodes

- Calculate the variance in action distributions between episodes

- Note any evolution or change in behavior across sequential episodes

- Determine if the agent displays consistent preferences or purely situational responses

\noindent Achievement Patterns

- Calculate achievement unlock rate and consistency across episodes

- Identify which achievements are most frequently obtained

- Analyze the typical sequence or prerequisites leading to achievements

- Assess whether achievement patterns suggest intentional pursuit or incidental acquisition

\noindent Resource Priorities

- Identify primary resources consistently targeted across episodes

- Analyze typical crafting sequences when resources are available

- Evaluate how consistently the agent manages inventory

- Determine if there are clear resource collection preferences

\noindent Environmental Interaction Patterns

- How consistently does the agent navigate the environment?

- Identify common responses to specific environmental features

- Analyze patterns in exploration vs. exploitation behavior

- Evaluate adaptation to threats, opportunities, and constraints

\noindent Decision-Making Characteristics

- Identify the apparent decision criteria for different action choices

- Analyze how the agent balances short-term vs. long-term needs

- Evaluate how predictable the agent's responses are to similar situations

- Assess whether actions appear purposeful or random

\noindent [Output Instructions]

1. Begin with a "Behavioral Profile" summarizing the agent's most consistent traits

2. Include a "Statistical Analysis" section with quantitative breakdowns of action patterns

3. Provide a "Decision Pattern Analysis" detailing how the agent makes choices

4. Add an "Achievement Analysis" showing typical patterns in achievement progression

5. Conclude with "Behavioral Consistency Assessment" that evaluates how predictable the agent is

\noindent Give the agent a label based on its observed behavior and justify your choice.
Your analysis should be based entirely on observable patterns. If the agent shows highly inconsistent behavior across episodes, explicitly detail this with supporting evidence. Focus on describing what the agent does consistently, rather than speculating on why it might do so.
\end{ttfamily}

\section{Implementation Details}
\label{apx:implementation_details}
\subsection{MiniGrid}
We employed the PPO algorithm from the stable-baselines3 library for our policy network, which takes as input a $K \times K \times 3$ encoded image and a mission string, the latter being encoded using a one-hot scheme. These inputs are combined into a single 2835-dimensional vector. The network architecture features two hidden layers, each comprising 64 neurons, with ReLU activation functions introducing non-linearity. The output layer, designed to match the 6-dimensional action space of the environment, utilizes a softmax activation function to generate a probability distribution over possible actions. Additionally, we normalized the observations. For the short-sighted agent, the observation grid size is $3 \times 3 \times 3$, while for the goal-directed agent, it is $11 \times 11 \times 3$.

\begin{table}[htbp]
    \centering
    \resizebox{\columnwidth}{!}{ 
    \begin{tabular}{lcccccc}
        \hline
        Hyperparameter & Goal-Directed & Short-Sighted & Dynamic Obstacles & Lava Gap & Red Blue Doors & Crossing \\
        \hline
        Total Timesteps          & $2 \times 10^6$  & $1 \times 10^6$  & $2 \times 10^6$  & $2 \times 10^6$  & $2 \times 10^6$  & $3 \times 10^6$ \\
        Number of Environments   & 8                & 8                & 8                & 16                & 8                & 16                \\
        Number of Steps          & 512              & 512              & 2048              & 1024              & 512              & 2048              \\
        Batch Size               & 64               & 64               & 256               & 128               & 64               & 256               \\
        GAE Lambda (\texttt{gae\_lambda}) & 0.95     & 0.95             & 0.95             & 0.95             & 0.95             & 0.95             \\
        Discount Factor (\texttt{gamma})  & 0.99     & 0.99             & 0.99             & 0.99             & 0.99             & 0.99             \\
        Number of Epochs         & 10               & 10               & 30               & 10               & 10               & 20               \\
        Entropy Coefficient      & 0.001            & 0.001            & 0.01            & 0.001            & 0.001            & 0.01            \\
        Learning Rate            & $1 \times 10^{-4}$ & $1 \times 10^{-4}$ & $1 \times 10^{-4}$ & $1 \times 10^{-4}$ & $1 \times 10^{-4}$ & $1 \times 10^{-4}$ \\
        Clip Range               & 0.2              & 0.2              & 0.2              & 0.2              & 0.2              & 0.2              \\
        \hline
    \end{tabular}
    }
    \caption{Hyper-parameters for the PPO algorithm applied to all six agents.}
    \label{tab:hyperparameters_all_agents}
\end{table}

\subsection{Crafter}
We implemented DreamerV3 for our agent, using a state-of-the-art world model-based reinforcement learning approach. The agent processes $64 \times 64 \times 3$ RGB observations from the Crafter environment. The world model consists of three key components: an encoder network, a recurrent state-space model (RSSM), and a decoder network. The encoder transforms raw pixel observations into a 1024-dimensional embedding space using a convolutional neural network with a depth of 96 channels.

The RSSM, which forms the core of the agent's predictive capabilities, utilizes a deterministic state of dimension 4096 and a stochastic state represented as a 32-dimensional random variable, allowing the agent to account for environment stochasticity. For temporal dynamics, we employed a GRU cell with 1024 hidden units. The decoder reconstructs observations using transposed convolutions, enabling the model to learn compact state representations through reconstruction loss.

For policy learning, we used an actor-critic architecture with 5-layer MLPs for both actor and critic, where the actor employs a categorical distribution over the 17 discrete actions available in Crafter. The agent was trained using the "reinforce" gradient strategy for imagination-based policy optimization, with a $\lambda$-return horizon of 15 steps and a discount factor of 0.997.

Training was conducted for $10^6$ environment steps using 8 parallel environments, with a batch size of 32 and sequence length of 64. We employed a model learning rate of $10^{-4}$ and an actor learning rate of $3 \times 10^{-5}$, optimized using Adam.
\section{Experts instructions:}
\label{apx:experts_instructions}
\noindent
\textbf{General Instructions:}\\
Generate a focused summary of the RL agent’s policy based on the provided episodes data. Highlight key behaviors, decision-making processes, and patterns specific to this agent. Tailor the summary to reflect unique strategies and actions observed.


\noindent\textbf{Focus on:}
\begin{itemize}
    \item Recurring patterns and behaviors specific to this agent’s policy.
    \item Detailed analysis of decision-making processes and responses to different stimuli.
    \item Efficiency in identifying and interacting with relevant objects (e.g., keys, doors).
    \item Methods used to solve tasks and handle obstacles.
    \item Comparison of agent’s performance across different episodes.
    \item Quantitative metrics (e.g., number of steps, success rates) to evaluate efficiency.
    \item Analysis of navigation strategies and adaptations to the environment.
    \item Provide insights on the variability and randomness of the agent’s actions and decisions. Look at the distribution of the agent’s actions during the episodes.
    \item Describe any notable fluctuations or inconsistencies in behavior across episodes.
    \item Figure out what is the agent’s field of vision and its influence on its decisions and actions, considering it may not see the entire grid but only a part of it.
\end{itemize}


\noindent\textbf{Environment Description:}
\begin{itemize}
    \item Goal: Maximize cumulative reward (close to 1) by opening the door.
    \item The environment is static, with only the agent moving within it. Walls are fixed and form the boundaries of the grid (there are no further obstacles).
    \item Possible agent’s actions: move forward, turn left, turn right, pick up, toggle, drop.
    \item Max episode length: 70 steps (failure to open the door if exceeded).
\end{itemize}


\noindent\textbf{Summary Instructions:}\\
The agent description should be at least 100 words. Provide approximately 5 key insights.

\section{Scale Used in Task 1}
\label{apx:task1_scale}
In Task 1 of our study, we utilized a 7-point Likert scale to evaluate participants' perceptions and understanding of the agent's behavior as presented in both the video summaries and the natural language summaries. Participants rated their agreement with the following statements, where 1 indicates ``Strongly disagree'' and 7 indicates ``Strongly agree''. The questions were phrased according to the condition—either video or natural language summary.

\begin{enumerate}
    \item \textbf{Clarity}: ``The [video/natural language] summary clearly explained the agent’s actions and decisions shown in the demonstration video.''
    \item \textbf{Understandable}: ``From the [video/natural language] summary, I understand how the agent’s actions and decisions shown in the demonstration video.''
    \item \textbf{Completeness}: ``The [video/natural language] summary seemed complete in covering all aspects of the agent’s actions and decisions in the demonstration video.''
    \item \textbf{Satisfaction}: ``The [video/natural language] summary is satisfying in capturing the agent’s behavior and decisions displayed in the demonstration video.''
    \item \textbf{Useful}: ``The [video/natural language] summary is useful to my understanding of the agent’s behavior and decisions displayed in the demonstration video.''
    \item \textbf{Accuracy}: ``The information in the [video/natural language] summary accurately reflected the agent’s behavior and decisions displayed in the demonstration video.''
    \item \textbf{Improvement}: ``The [video/natural language] summary provides additional insights about the agent’s behavior that are not immediately apparent from watching the demonstration video alone.''
    \item \textbf{Preference}: ``I prefer receiving information about agent behavior through the [video/natural language] summary rather than just watching the demonstration video.''
\end{enumerate}

These ratings provided quantitative data to assess the effectiveness and clarity of both the video and natural language summaries in conveying the agent’s behavior and decision-making processes. This scale aimed to capture various dimensions of participant satisfaction and understanding, contributing to the overall evaluation of the summaries' utility in the context of our research.

\section{Systematic Exploration of the Prompt Design}
\label{apx:prompt_design}
The creation of the final prompt was achieved through a structured and iterative exploration process. This process involved a quantitative evaluation of prompt designs based on observed outputs, guided by principles from prompt engineering literature, and tailored to domain-specific requirements. Additionally, the final design was inspired by the Chain of Thought (CoT)~\cite{wei2022chain} prompting paradigm, which encourages models to generate structured, step-by-step reasoning. Below is a detailed breakdown of the methodology used:

\subsection*{Define the Objective}

\textbf{Goal:} The primary objective of the prompt was to generate a focused and comprehensive global summary of the policy of the RL agent. The summary needed to highlight key behaviors, decision-making processes, and performance metrics in a manner understandable to both technical and non-technical audiences, while ensuring it could function as a \textit{zero-shot prompt} without requiring additional training examples.

\textbf{Key Constraints:}

\begin{itemize}
    \item The prompt must guide the model to produce specific, concise, and informative summaries.
    \item It should minimize general or trivial statements and focus on insights from the agent’s behavior.
\end{itemize}

\subsection*{Decomposition of Requirements}

To meet the objective, the task was broken down into several core components:
\begin{itemize}
    \item \textbf{Behavioral Analysis:} Capturing recurring patterns, strategies, and responses to stimuli.
    \item \textbf{Performance Metrics:} Including quantitative insights such as success rates and steps taken.
    \item \textbf{Environmental Factors:} Reflecting the influence of the agent’s field of vision and static surroundings.
    \item \textbf{Comparison Across Episodes:} Addressing variability and randomness in actions.
    \item \textbf{Accessibility:} Ensuring the output is clear and digestible for non-technical readers.
\end{itemize}

\subsection*{Iterative Prompt Design}

\textbf{Initial Prototype:}
\begin{itemize}
    \item Focused on general instructions for summarization.
    \item Included high-level tasks such as ``describe the agent's behavior'' without specifying details.
\end{itemize}

\textbf{Issues Identified:}
\begin{itemize}
    \item Outputs were overly generic, lacked depth, and failed to focus on specific behaviors or metrics.
\end{itemize}

\textbf{Refinement 1: Add Specific Focus Areas}
\begin{itemize}
    \item Incorporated bullet points to guide the model to focus on particular aspects, such as ``recurring patterns,'' ``quantitative metrics,'' and ``navigation strategies.''
\end{itemize}

\textbf{Observations:}
\begin{itemize}
    \item Improved relevance and depth of the summaries.
    \item However, the outputs lacked consistency in formatting and interpretability.
\end{itemize}

\textbf{Refinement 2: Structured Prompt Sections}
\begin{itemize}
    \item Segmented the prompt into distinct parts:
    \begin{itemize}
        \item General Instructions
        \item Environment Description
        \item Interpretation Instructions
        \item Output Instructions
        \item Formatted Experience Dataset
    \end{itemize}
\end{itemize}

\textbf{Observations:}
\begin{itemize}
    \item Enhanced structure improved consistency.
    \item More detailed context in ``Environment Description'' provided clarity for the model to ground its responses.
\end{itemize}

\textbf{Refinement 3: Inspired by Chain of Thought (CoT) Reasoning}
\begin{itemize}
    \item The prompt was designed to encourage a step-by-step analysis, mirroring the CoT paradigm:
    \begin{itemize}
        \item Each bullet point and section was treated as a sub-task requiring focused attention.
        \item For example, instructions like ``Analyze navigation strategies and adaptations to the environment'' explicitly directed the model to break down its reasoning into smaller, manageable steps.
    \end{itemize}
\end{itemize}

\textbf{Observations:}
\begin{itemize}
    \item Outputs exhibited improved logical flow and comprehensive coverage of required aspects.
    \item The structured approach mitigated issues with overly generic or shallow responses.
\end{itemize}

\textbf{Refinement 4: Emphasize Quantitative and Comparative Analysis}
\begin{itemize}
    \item Added explicit instructions to include metrics like ``number of steps'' and ``success rates.''
    \item Introduced the requirement to compare the agent's performance across episodes.
\end{itemize}

\textbf{Observations:}
\begin{itemize}
    \item Summaries became more data-driven and analytical.
    \item Increased attention to variations in the agent’s behavior.
\end{itemize}

\textbf{Refinement 5: Addressing Accessibility}
\begin{itemize}
    \item Adjusted language in the ``Output Instructions'' to ensure summaries were understandable to non-technical audiences.
    \item Included a directive to avoid trivial statements.
\end{itemize}

\textbf{Final Testing:}
\begin{itemize}
    \item Conducted multiple test runs with varied episode datasets.
    \item Evaluated the prompt’s ability to guide the model toward producing outputs that met the objective.
    \item Fine-tuned phrasing for clarity and focus.
\end{itemize}

\subsection*{Key Design Considerations}

\textbf{Clarity and Specificity:}
\begin{itemize}
    \item Each section of the prompt was crafted to minimize ambiguity, ensuring the model understood the task requirements.
\end{itemize}

\textbf{Structure Inspired by CoT:}
\begin{itemize}
    \item The step-by-step breakdown mirrored the CoT prompting approach, which is known to improve reasoning and response quality in large language models.
\end{itemize}

\textbf{Focus on Insightful Analysis:}
\begin{itemize}
    \item By explicitly asking for ``variability,'' ``distribution of actions,'' and ``quantitative comparisons,'' the prompt steered the model toward generating meaningful insights.
\end{itemize}

\subsection*{Evaluation and Lessons Learned}

\textbf{Evaluation:}
\begin{itemize}
    \item Outputs were analyzed for relevance, specificity, and clarity.
    \item Feedback from test runs informed iterative improvements.
\end{itemize}

\textbf{Lessons Learned:}
\begin{itemize}
    \item Prompts benefit from structured sections that provide clear and detailed guidance.
    \item Incorporating CoT-inspired design principles encourages logical, step-by-step reasoning in outputs.
    \item Tailoring language for accessibility improves utility for non-technical audiences.
\end{itemize}

\subsection*{Rationale for the Final Design}

The final prompt integrates the following elements:
\begin{itemize}
    \item \textbf{Comprehensive Instructions:} Ensuring detailed and targeted outputs.
    \item \textbf{Quantitative Focus:} Providing measurable insights for evaluating agent performance.
    \item \textbf{Clarity and Accessibility:} Catering to a broad audience, including non-technical users.
    \item \textbf{Structure Inspired by CoT:} Encouraging the model to follow a logical sequence in generating summaries.
\end{itemize}

This systematic process, incorporating insights from the Chain of Thought paradigm, demonstrates the thoughtful process taken to ensure the prompt is both effective and robust for summarizing RL agent policies.


\end{document}